%% file: paper.tex
\documentclass[]{fairmeta}
\usepackage{lineno}
\usepackage{calc}
\usepackage{amsmath}
\usepackage{caption}
\usepackage{multirow}
\usepackage{amsfonts}
\usepackage{xspace}
\usepackage[font=footnotesize]{subcaption}
\usepackage{booktabs}
\usepackage{pdfpages}
\usepackage{xcolor}
\usepackage{tabularx}
\usepackage{bbm}
\usepackage{graphicx}
\usepackage{algorithm}
\usepackage{algorithmic}
\usepackage{enumitem}
\usepackage{pifont}

\usepackage[export]{adjustbox}

\newcommand{\ours}{\textsc{MoFA}\xspace}

\title{LLM-Driven Reasoning for Constraint-Aware Feature Selection in Industrial Systems}

\author{Yuhang Zhou}
\author{Zhuokai Zhao}
\author{Ke Li}
\author{Spilios Evmorfos}
\author{Gökalp Demirci}
\author{Mingyi Wang}
\author{Qiao Liu}
\author{Qifei Wang}
\author{Serena Li}
\author{Weiwei Li}
\author{Tingting Wang}
\author{Mingze Gao}
\author{Gedi Zhou}
\author{Abhishek Kumar}
\author{Xiangjun Fan}
\author{Lizhu Zhang}
\author{Jiayi Liu}

\affiliation{Meta AI}

\abstract{\input{arxiv/0_abstract}}

\date{\today}
\correspondence{Yuhang Zhou, Jiayi Liu at \email{\{zyhang, liujiayi\}@meta.com}\\}

\begin{document}

\maketitle

\input{arxiv/1_introduction}
\input{arxiv/2_related}

\input{arxiv/3_method}
\input{arxiv/5_experiments}
\input{arxiv/6_ablations}
\input{arxiv/7_conclusion}

\clearpage
\newpage
\bibliographystyle{assets/plainnat}
\bibliography{custom}

\clearpage
\newpage
\beginappendix
\input{arxiv/8_appendix}

\end{document}

%% file: arxiv/0_abstract.tex
Feature selection is a crucial step in large-scale industrial machine learning systems, directly affecting model accuracy, efficiency, and maintainability. Traditional feature selection methods rely on labeled data and statistical heuristics, making them difficult to apply in production environments where labeled data are limited and multiple operational constraints must be satisfied. To address this, we propose \textbf{Model Feature Agent} (\ours), a large language model–driven framework that performs sequential, reasoning-based feature selection using both semantic and quantitative feature information. \ours\ incorporates feature definitions, importance scores, correlations, and metadata (e.g., feature groups or types) into structured prompts and selects features through interpretable, constraint-aware reasoning. We evaluate \ours\ in three real-world industrial applications: (1) True Interest and Time-Worthiness Prediction, where it improves accuracy while reducing feature group complexity; (2) Value Model Enhancement, where it discovers high-order interaction terms that yield substantial engagement gains in online experiments; and (3) Notification Behavior Prediction, where it selects compact, high-value feature subsets that improve both model accuracy and inference efficiency. Together, these results demonstrate the practicality and effectiveness of LLM-based reasoning for feature selection in real production systems.

%% file: arxiv/1_introduction.tex
\section{Introduction}
\label{sec:intro}

Feature selection is a fundamental component of industrial machine learning systems, directly influencing model accuracy, interpretability, and efficiency \citep{li2017feature, kumar2014feature, chandrashekar2014survey}. In large-scale production environments such as recommendation and notification systems, models often rely on hundreds of heterogeneous features collected from user behavior, content attributes, and system signals \citep{eirinaki2018recommender}. Selecting the right subset of features is therefore essential for both predictive performance and maintainability. However, traditional feature selection approaches depend heavily on labeled training data and numerical statistics \citep{lemhadri2021lassonet, tibshirani1996regression}, making them difficult to apply when labeled data are scarce or when additional business or operational requirements must be satisfied.

Recent advances in large language models (LLMs) provide new opportunities for feature reasoning. LLMs are capable of understanding both the semantic meaning and quantitative context of features described in natural language \citep{jeong2024llm, choi2022lmpriors, li2024improving}. This capability opens a new paradigm for feature selection that extends beyond purely statistical criteria to incorporate contextual and domain-specific knowledge. However, existing LLM-based approaches focus solely on optimizing predictive utility while overlooking practical constraints, such as capacity limits or maintainability, and often fail to leverage the rich metadata naturally available in production systems. These limitations motivate a more general and constraint-aware framework for LLM-driven feature selection in real-world industrial settings.

In this work, we propose Model Feature Agent (\ours), an LLM-based framework for feature selection in industrial systems. \ours formulates feature selection as a sequential reasoning process: at each step, the LLM receives structured information about the task objective, model configuration, current and available features, and auxiliary requirements, and then selects the next most suitable feature along with interpretable reasoning. To address the scalability challenges in industrial settings with thousands of features, we introduce a Divide-and-Conquer strategy that partitions the feature space into manageable buckets for parallel selection followed by a global refinement phase.

We evaluate \ours\ through three real-world industrial use cases. In the True Interest and Time-Worthiness Prediction task, \ours\ improves accuracy while reducing feature group complexity. In the Value Model Enhancement task, it identifies meaningful higher-order interaction terms, and online experiments show that incorporating these terms into the value model yields substantial gains in key engagement metrics. In the Notification Behavior Prediction task, \ours\ selects compact, high-value feature subsets from a large heterogeneous feature space, improving both model accuracy and inference efficiency. Together, these results demonstrate that LLM-based reasoning can serve as a practical and effective tool for feature selection in modern production systems, bridging the gap between interpretability, flexibility, and real-world scalability.

%% file: arxiv/2_related.tex
\section{Related Work}
\label{sec:related}

\paragraph{Feature Selection with Classical Models}
Feature selection aims to identify the most informative subset of features that improves model generalization, interpretability, and efficiency \citep{blum1997selection, tang2014feature, dash1997feature, guyon2003introduction}. Existing methods are commonly categorized into three groups: \textbf{filter}, \textbf{wrapper}, and \textbf{embedded} approaches. Filter methods evaluate feature relevance independently of any specific model, typically using statistical measures such as correlation, mutual information, or variance scores \citep{lazar2012survey, cherrington2019feature, dash2002feature, lewis1992feature, ding2005minimum}. Wrapper methods, in contrast, rely on iterative model training and evaluation to assess feature subsets \citep{kohavi1997wrappers, chen2015novel, uncu2007novel, yamada2020feature}, which often yields higher accuracy but incurs substantial computational cost, such as sequential selection and recursive feature elimination \citep{ferri1994comparative, luo2014sequential, guyon2003introduction}. Embedded methods integrate feature selection directly into the model training process, for example through sparsity-inducing regularization or tree-based importance measures \citep{tibshirani1996regression, yuan2006model, feng2017sparse}. Despite their effectiveness in data-rich scenarios, these traditional approaches require access to sufficient labeled data and cannot leverage the semantic or metadata information associated with each feature, limitations that motivate our use of large language models for more context-aware and flexible feature selection.

\paragraph{Feature Selection with LLMs}

Recent studies have explored using LLMs to assist or automate feature selection through natural language reasoning and prompt-based evaluation. One line of work formulates feature selection as a classification task, where the LLM is prompted to determine whether a given feature is important for a specific prediction goal or to assign an importance score based on textual descriptions \citep{jeong2024llm, choi2022lmpriors}. Another line of research treats it as a ranking problem, leveraging the LLM’s ability to re-rank features according to their relevance or contribution to model performance \citep{jeong2024llm, li2025exploring, li2025llm4fs}. A third direction refines feature selection iteratively, incorporating feedback from downstream model results to improve subsequent selections \citep{han2024large, yang2024ice}. However, these approaches typically rely on labeled data and controlled settings, whereas our work targets real-world industrial scenarios with limited supervision, rich feature metadata, and multiple operational constraints.

%% file: arxiv/3_method.tex
\section{\ours Workflow}
\label{sec:workflow}

Feature selection plays a crucial role in industrial recommender systems, where models must balance multiple goals such as predictive accuracy, user engagement, and system efficiency. Traditional approaches mainly rely on numerical correlations or model-based importance scores, often ignoring the semantic and contextual meaning of features. As a result, these methods may fail to capture how different features relate to product objectives or operational constraints.

\begin{figure}
    \centering
    \includegraphics[width=0.8\linewidth]{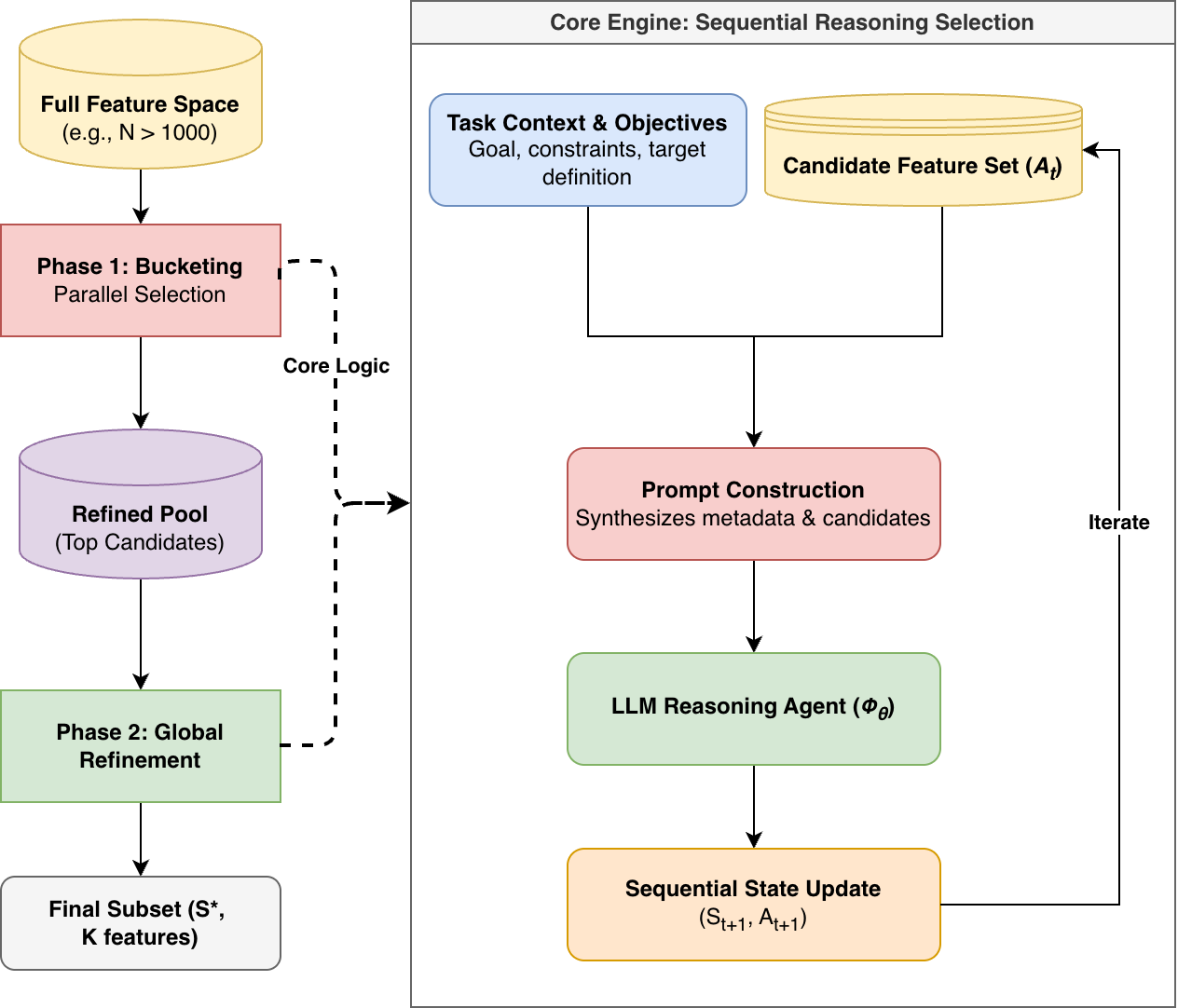}
    \caption{The \textbf{\ours} pipeline with Divide-and-Conquer scaling. The process begins by partitioning the candidate pool into disjoint buckets (Phase 1). An LLM agent ($\Phi_\theta$) performing sequential reasoning first selects local candidate subsets from each bucket. These subsets are merged into a refined candidate pool, where the agent performs a final round of global sequential selection (Phase 2) to reach the target size $K$.}
    \label{fig:agent_workflow}
\end{figure}

To address this gap, we propose the \textbf{\ours} pipeline. As shown in Figure \ref{fig:agent_workflow}, \ours leverages LLMs as reasoning agents that can analyze not only the descriptions of features but also their associated metadata, such as feature importance and correlations with key metrics. By integrating these heterogeneous sources of information, \ours performs sequential, context-aware feature selection that aims to improve model performance while satisfying additional business requirements.

\subsection{Problem Formulation}

Let $\mathcal{F} = \{f_1, f_2, \dots, f_N\}$ denote the set of available features, and let $\mathcal{S}_t \subseteq \mathcal{F}$ denote the current feature subset selected at step $t$. Each feature $f_i$ is represented as a tuple: $f_i = \big( \text{name}_i, \text{desc}_i, \text{meta}_i \big)$, where $\text{name}_i$ and $\text{desc}_i$ are the textual name and description of the feature, and $\text{meta}_i$ contains auxiliary quantitative information (e.g., feature importance scores, correlation coefficients with target metrics, or group constraints).

Given a target objective function $\mathcal{O}$ (e.g., model performance, engagement lift, or a weighted combination of business metrics), the goal is to select a subset $\mathcal{S}^* \subseteq \mathcal{F}$ satisfying:
\[
\mathcal{S}^* = \arg\max_{\mathcal{S} \subseteq \mathcal{F}} \mathcal{O}(\mathcal{S}) \quad \text{s.t.} \quad \mathcal{C}(\mathcal{S}) = \text{true},
\]
where $\mathcal{C}(\mathcal{S})$ represents a set of auxiliary constraints such as capacity limits, feature group restrictions, or fairness requirements.

\subsection{Scalable Sequential Selection via Divide-and-Conquer}

In industrial settings where the total number of features $N$ exceeds the context window of the LLM, we adopt a two-phase \textbf{Divide-and-Conquer} approach. Both phases utilize the same core sequential reasoning selection logic to maintain consistency and depth.

\paragraph{Phase 1: Bucketed Parallel Selection}
We first partition the total feature set $\mathcal{F}$ into $B$ disjoint buckets $\{\mathcal{B}_1, \dots, \mathcal{B}_B\}$. For each bucket $\mathcal{B}_b$, the \ours agent performs sequential selection until it identifies a candidate subset $\mathcal{S}_b$ where $|\mathcal{S}_b| = K' > K/B$. This ensures that highly relevant features from each partition are preserved for global evaluation while staying within the model's context window during each reasoning step.

\paragraph{Phase 2: Global Synthesis and Refinement}
The subsets are merged into a unified candidate pool $\mathcal{F}_{refine} = \bigcup_{b=1}^{B} \mathcal{S}_b$. The agent then performs the final round of sequential reasoning selection on $\mathcal{F}_{refine}$ to reach the target size $K$. By evaluating the top candidates from all buckets simultaneously, the LLM can reason about dependencies and redundancies across the entire filtered feature space, mitigating collinearity problems that might otherwise arise from independent bucket selections.

\subsection{Sequential Selection via LLM Reasoning}

The \ours agent decomposes the selection process into a series of LLM-driven reasoning steps, where one feature is selected at each iteration. At step $t$, given the current feature set $\mathcal{S}_t$ and the remaining available features $\mathcal{A}_t$ (within a specific bucket or the refined pool), the LLM receives a structured prompt $\mathcal{P}_t$ that includes: (1) A description of the task and model objective; (2) The list of features currently in use, $\mathcal{S}_t$, along with their associated metadata; (3) The list of candidate features, $\mathcal{A}_t$, each annotated with semantic information; and (4) Optional auxiliary requirements $\mathcal{C(S)}$.

Formally, the LLM is prompted to output:
\[
f^*_t = \arg\max_{f_i \in \mathcal{A}_t} \Phi_\theta(f_i \mid \mathcal{S}_t, \mathcal{O}, \mathcal{R}),
\]
where $\Phi_\theta$ denotes the LLM’s internal reasoning and decision function parameterized by $\theta$. The selected feature $f^*_t$ is then added to the current set:
\[
\mathcal{S}_{t+1} = \mathcal{S}_t \cup \{f^*_t\}, \quad \mathcal{A}_{t+1} = \mathcal{A}_t \setminus \{f^*_t\}.
\]
The process continues until the required number of features for that specific phase is reached.

\paragraph{Prompt Construction and Output Format} 
At each selection step, \ours constructs a structured user prompt providing all necessary contextual information for informed decision-making. The LLM then reasons over these inputs to identify the next most suitable feature. Each reasoning step is decomposed into two parts: (1) the model’s natural language reasoning process, and (2) a structured selection output. The expected response format is:
\begin{quote}
\texttt{Selected Feature:} $f^*_t$, \texttt{Reason:} $\rho_t$
\end{quote}
where $\rho_t$ represents the explanation for why the selected feature $f^*_t$ best satisfies the main objective $\mathcal{O}$ while adhering to the auxiliary constraints $\mathcal{C}$.

%% file: arxiv/5_experiments.tex
\section{Experiments}
\label{sec:experiments}

We evaluate \ours in three real-world industrial applications, covering different model architectures and feature domains, to assess its effectiveness and adaptability across practical settings.

\subsection{Experimental Setup}

\paragraph{Models and Baselines.} 
For feature selection, we employ the \texttt{Llama4-Maverick} model as the backbone of the \ours\ agent \citep{meta2025llama}. This model serves as the reasoning engine that sequentially evaluates and selects features based on the constructed prompts. To provide meaningful comparisons, we include two types of baselines. For use cases with available offline training data, we adopt \texttt{Lasso} regression as a representative statistical feature selection method that leverages data-driven sparsity \citep{tibshirani1996regression}. For use cases without offline data, common in real-world recommendation systems where only online signals are available, we construct a baseline by randomly selected features in online experiments. This setup allows us to assess the performance gain of \ours\ under both data-rich and data-scarce environments.

%% file: arxiv/6_ablations.tex

\subsection{Prediction of True Interest and Time-Worthiness}

\paragraph{Background and Task}
In this use case, we seek to predict whether a piece of content (post or short video) reflects a user's \emph{true interest} and whether it is \emph{worth the user's time}. To obtain supervision aligned with user sentiment, we randomly sample content exposures and trigger an in-product survey that asks users to judge (i) whether the item reflects their true interest and (ii) whether it was worth their time. Let $y^{\mathrm{TI}} \in \{0,1\}$ and $y^{\mathrm{WT}} \in \{0,1\}$ denote the corresponding binary labels. Each training instance $i$ consists of a feature vector $x_i \in \mathbb{R}^{N}$ that concatenates user-level and content-level features, $x_i = [x^{(u)}_i; x^{(c)}_i]$, and the labels $\big(y^{\mathrm{TI}}_i, y^{\mathrm{WT}}_i\big)$. The learned predictors are integrated into the recommendation stack as additional signals (e.g., ranking features) to improve downstream relevance.

\paragraph{Modeling and Feature-Group Requirement}
We train logistic regression models for each label using a subset of features selected by \ours. Concretely, for task $q \in \{\mathrm{TI}, \mathrm{WT}\}$ we estimate $p^{(q)}(x; \beta^{(q)}) \;=\; \sigma\!\big((\beta^{(q)})^{\top} x_{S^{(q)}}\big),$ 
where $\sigma(\cdot)$ is the logistic link and $x_{S^{(q)}}$ indexes the coordinates of $x$ selected for task $q$. Features are partitioned into operational \emph{groups} $\mathcal{G}=\{G_1,\dots,G_G\}$ (e.g., by owning team or data pipeline), and our auxiliary requirement is to \emph{concentrate} the selected features into fewer groups to reduce cross-team dependencies and maintenance overhead. 


Within \ours, this auxiliary requirement is implemented by providing each candidate feature’s group identifier as part of its metadata ($\text{meta}_i$) in the prompt. During prompt construction, we explicitly instruct the LLM to prioritize selections that maintain fewer feature groups whenever possible. In this way, the LLM integrates the group information into its reasoning process, balancing predictive performance with operational simplicity. The resulting sequential, LLM-guided selection produces compact and maintainable feature sets that align well with practical engineering constraints.

\paragraph{Offline Experiment Setup} 
We evaluate \ours on two prediction tasks derived from user survey data: \emph{true interest} (TI) and \emph{time-worthiness} (WT). The TI and WT dataset comprises 71,754 training instances and 17,939 test instances. Both tasks predict binary labels based on user responses to in-product surveys. The feature space consists of user-level and content-level features organized into operational groups maintained by different teams. We evaluate feature subsets of size $K \in \{20, 100, 200, 500\}$ from a total of 1,030 features to assess scalability across sparse and dense feature regimes. To manage this feature space within the LLM's context window, we employ the Divide-and-Conquer strategy by partitioning the 1,030 features into $B=5$ disjoint buckets. In Phase 1, the agent selects $K' = 1.5 \times (K/B)$ features from each bucket. In Phase 2, the agent performs a final sequential selection to reach the target $K$. For the Lasso baseline, we use $L_1$-regularized logistic regression with regularization parameter $C=0.1$, selecting the top-$K$ features ranked by absolute coefficient magnitude.

Table~\ref{tab:results} presents the performance and feature group utilization for both methods across different feature budget sizes for both prediction tasks.

\begin{table}[t]
\centering
\small
\begin{minipage}[t]{0.48\textwidth}
    \centering
    \begin{tabular}{lcccc}
    \toprule
    \multicolumn{5}{c}{\textbf{True Interest}} \\
    \midrule
    \textbf{K} & \textbf{Method} & \textbf{Groups} & \textbf{Train} & \textbf{Test} \\
    \midrule
    \multirow{2}{*}{20} 
    & Lasso & 9 & 74.49 & 71.51 \\
    & \ours & \textbf{13} & \textbf{75.18} & \textbf{72.14} \\
    \midrule
    \multirow{2}{*}{100} 
    & Lasso & 48 & 76.98 & 73.77 \\
    & \ours & \textbf{45} & \textbf{77.04} & \textbf{73.93} \\
    \midrule
    \multirow{2}{*}{200} 
    & Lasso & 87 & \textbf{77.86} & \textbf{74.58} \\
    & \ours & \textbf{61} & 77.87 & 74.50 \\
    \midrule
    \multirow{2}{*}{500} 
    & Lasso & 187 & \textbf{78.93} & \textbf{75.22} \\
    & \ours & \textbf{85} & 78.68 & 74.81 \\
    \bottomrule
    \end{tabular}
\end{minipage}
\hfill
\begin{minipage}[t]{0.48\textwidth}
    \centering
    \begin{tabular}{lcccc}
    \toprule
    \multicolumn{5}{c}{\textbf{Time-Worthiness}} \\
    \midrule
    \textbf{K} & \textbf{Method} & \textbf{Groups} & \textbf{Train} & \textbf{Test} \\
    \midrule
    \multirow{2}{*}{20} 
    & Lasso & 19 & 69.29 & \textbf{69.72} \\
    & \ours & \textbf{15} & \textbf{69.11} & 69.69 \\
    \midrule
    \multirow{2}{*}{100} 
    & Lasso & 84 & \textbf{73.30} & \textbf{72.86} \\
    & \ours & \textbf{54} & 72.92 & 72.66 \\
    \midrule
    \multirow{2}{*}{200} 
    & Lasso & 84 & \textbf{74.22} & \textbf{73.76} \\
    & \ours & \textbf{76} & 73.74 & 73.44 \\
    \midrule
    \multirow{2}{*}{500} 
    & Lasso & 313 & \textbf{75.53} & \textbf{74.65} \\
    & \ours & \textbf{96} & 74.24 & 73.62 \\
    \bottomrule
    \end{tabular}
\end{minipage}
\caption{Prediction results for true interest and time-worthiness tasks. \textbf{Groups} denotes the number of distinct feature groups in the selected subset; Train and Test refer to AUC scores.}
\label{tab:results}
\end{table}

\paragraph{Predictive Accuracy} 
The relationship between \ours and Lasso varies by task and budget. For TI prediction, \ours outperforms Lasso at smaller budgets ($+0.63\%$ at $K=20$, $+0.16\%$ at $K=100$), demonstrating that semantic reasoning effectively identifies informative features under tight constraints. At larger budgets ($K \geq 200$), both methods converge to similar performance (within $0.5\%$ AUC). For WT prediction, Lasso achieves slightly higher test AUC across all budgets, though the gap is less than $0.5\%$ in most cases. Despite modest AUC differences, \ours delivers substantial operational benefits through feature group consolidation, as discussed below.

\paragraph{Feature Group Efficiency} 
The most significant advantage of \ours is its ability to dramatically reduce the number of feature groups required, thereby lowering cross-team coordination overhead and maintenance complexity. For TI prediction, \ours achieves a \textbf{30\% reduction} in groups at $K=200$ (61 vs 87) and a \textbf{54\% reduction} at $K=500$ (85 vs 187). For WT prediction, the gains are even more pronounced: a \textbf{10\% reduction} at $K=200$ (76 vs 84), and a dramatic \textbf{69\% reduction} at $K=500$ (96 vs 313 groups).

These results demonstrate that \ours's LLM-based reasoning successfully identifies features that balance predictive signal with operational alignment. While Lasso selects features purely based on statistical correlation, often choosing multiple features from different groups, \ours explicitly considers group membership as part of its decision process, concentrating selections within fewer groups. This is particularly valuable in industrial settings where each additional feature group introduces dependencies on different data pipelines, engineering teams, and maintenance workflows.

\subsection{Signal Pair Selection for Value Model}

\paragraph{Background and Task}
In large-scale recommendation systems, a \emph{value model} (VM) is commonly used to estimate the overall value of user interactions. The VM serves as a key optimization signal, helping balance engagement, satisfaction, and monetization. In our current production system, the value model is defined as a linear combination of major behavioral signals, such as:
$
\mathrm{VM} = a \cdot p_{\text{like}} + b \cdot p_{\text{click}},
$
where $p_{\text{like}}$ and $p_{\text{click}}$ denote the predicted probabilities of user liking and clicking on a piece of content, respectively. 

While this linear form is interpretable, it fails to capture interaction effects among behavioral signals, e.g., how simultaneous high values of both $p_{\text{like}}$ and $p_{\text{click}}$ may jointly reflect stronger user engagement. To better represent such dependencies, we consider extending the value model with higher-order interaction terms, such as:
$
\mathrm{VM}' = a \cdot p_{\text{like}} + b \cdot p_{\text{click}} + c \cdot (p_{\text{like}} \times p_{\text{click}}).
$
However, enumerating all possible signal pairs is computationally expensive in online environments. Therefore, we employ \ours\ to automatically identify the most promising interaction terms to include.

\paragraph{Feature Construction and Metadata}
We construct a candidate pool $\mathcal{P} = \{(s_i, s_j)\}$ of signal pairs, where each pair represents a potential multiplicative interaction term. For each candidate pair $(s_i, s_j)$, we compute and record the following metadata: (1) the individual correlations between $s_i$ and key engagement outcomes (e.g., user likes, saves, comments), (2) the pairwise correlation between $s_i \times s_j$ and these engagement signals, and (3) optional measures of signal reliability or coverage. These metadata fields are appended to the prompt as part of each candidate pair’s description, allowing the LLM to reason quantitatively about potential contribution and redundancy.

\paragraph{Online Experiment Results}

From a pool of approximately 400 signal pairs, \ours~selected the top-10 feature interactions, as shown in Table~\ref{tab:top_features} in Appendix. The three highest-ranked interactions: $p_\text{external share} \times p_\text{profile tap}$, $p_\text{reshare button tap} \times p_\text{profile tap}$, and $p_\text{reshare button tap} \times p_\text{comment}$, were sequentially incorporated into the value estimation model to assess their marginal contributions.

To evaluate the \ours approach, we measured the total number of daily app sessions across users. For each new feature integration, we assessed the corresponding lift in the session metric relative to the baseline production recommendation system. The session improvements were $0.055\%$, $0.048\%$, and $0.018\%$, all of which are statistically significant. These evaluations validate the effectiveness of the \ours~feature selection process: By surfacing high-value feature pairs, our method advances holistic engagement optimization and provides a scalable path for the continual improvement of large-scale recommender systems. 


\subsection{Notification Behavior Prediction}

\paragraph{Background and Task}
In large-scale mobile applications, personalized notification delivery plays a key role in maintaining user engagement while avoiding overexposure or fatigue. Our production system employs a multi-task learning framework to predict user behavior across a wide range of notification-related outcomes. The downstream model simultaneously generates $38$ prediction labels, including user click-through rate, dismissal rate, and other engagement signals. Among these, we primarily focus on improving two key metrics: the probability of \textbf{user click} and the probability of \textbf{user dismissing notifications}, as these directly capture user interest and annoyance balance. The production model integrates multiple sub-networks and architectural modules, including multi-layer perceptrons, gating networks, and aggregation mechanisms, to jointly process heterogeneous feature types. The combination of such diverse inputs yields a feature space that is both large and heterogeneous, making manual or brute-force feature selection infeasible.

\paragraph{Feature Construction and Metadata} 
For this case, we construct a comprehensive pool of candidate features drawn from the production model. In our setup, we use \ours~ to select 4,000 features from a total of 8,169 features. To effectively process this high-dimensional space, we partition the 8,169 features into $B=10$ disjoint buckets, ensuring each prompt remains within the LLM's context window. In Phase 1, the agent performs parallel reasoning to select 450 features from each bucket ($K' = 4,500$). In Phase 2, these candidates are merged into a refined pool where a final global selection pass is performed to reach the target of 4,000 features. Each feature is associated with metadata extracted from the currently deployed model. Specifically, for every candidate feature $f_i$, we include: its \textbf{feature category}, indicating whether it is a sparse, dense, or float feature. This metadata helps the LLM understand the modality of each feature. We also provide the LLM with the background of the task, including the model architecture and our main prediction objectives: user click probability and dismiss probability.


\paragraph{Experiment Results}

To evaluate the effectiveness of \ours in multi-task notification behavior prediction, we conducted offline experiments using normalized entropy (NE) loss as the primary metric on the test dataset. The current production model utilizes over 7,500 features, while our experiments randomly sampled 4,000 features for comparison. We report results for two key prediction tasks: probability of user click ($p_\text{click}$) and probability of user dismissing notifications ($p_\text{dismiss}$).

With 4,000 features selected by \ours, the evaluation NE loss shows relative improvements over the production model (over 7,500 features). Specifically, $p_\text{click}$ achieves a 0.065\% NE win, and $p_\text{dismiss}$ achieves a 0.198\% NE win compared to the current production setup. Compared to a baseline where 4,000 features were selected via random shuffling, \ours~achieves a 0.19\% NE win for $p_\text{click}$ and a 0.332\% NE win for $p_\text{dismiss}$.
These results demonstrate that \ours pipeline yields measurable improvements in notification behavior prediction, outperforming both the production model and random feature selection baselines.

%% file: arxiv/7_conclusion.tex
\section{Conclusion}
\label{sec:conclusion}

In this work, we propose \ours, an LLM-driven feature selection framework that integrates semantic understanding and quantitative metadata to enable context-aware and constraint-satisfying feature selection. Across real-world industrial cases, \ours\ consistently improves downstream performance and operational efficiency: enhancing prediction accuracy and maintainability in user interest modeling, identifying meaningful interaction signals, and selecting high-utility feature sets for multi-task notification prediction. These results demonstrate the practicality and versatility of leveraging LLM reasoning for intelligent feature selection in complex production environments.

\section{Limitations}

\ours framework offers a sophisticated paradigm shift by treating feature selection as a sequential reasoning process. By integrating semantic descriptions with quantitative metadata, Though \ours achieves high predictive utility while satisfying complex operational constraints. several limitations remains.

First, because \ours selects features one-by-one through iterative LLM prompting , the computational time scales linearly with the number of features $K$ to be selected. This may introduce latency in systems requiring thousands of features. However, this is mitigated by the fact that feature selection is typically an offline preprocessing step, making reasoning time less critical than the resulting model efficiency.

Second, in large-scale industrial systems with thousands of candidates, providing a comprehensive list of all features and their metadata (e.g., importance scores, correlations) can challenge the input token limits of the LLM. 

Finally, \ours's sequential approach is inherently a greedy search, which might overlook optimal feature combinations that only show synergy when selected together.

We leave these refinements of current pipeline to the future work to broaden the applicability of \ours pipeline.

%% file: arxiv/8_appendix.tex
\appendix
\section*{Appendix}

\begin{table}[h]
\centering
\begin{tabular}{cl}
\hline
\textbf{Rank} & \textbf{Feature Description} \\
\hline
1 & $p_\text{external share} \times p_\text{profile tap}$ \\
2 & $p_\text{reshare button tap} \times p_\text{profile tap}$ \\
3 & $p_\text{reshare button tap} \times p_\text{comment}$ \\
4 & $p_\text{reshare button tap} \times p_\text{skip}$ \\
5 & $p_\text{external share} \times p_\text{skip}$ \\
6 & $p_\text{reshare button tap} \times p_\text{log time}$ \\
7 & $p_\text{reshare button tap} \times p_\text{skip}$ \\
8 & $p_\text{external share} \times p_\text{log time}$ \\
9 & $p_\text{comment surface enter} \times p_\text{external share}$ \\
10 & $p_\text{like} \times p_\text{reshare button tap}$ \\
\hline
\end{tabular}
\caption{Top-10 feature interactions selected by \ours. Each feature represents the product of two engagement signal probabilities. The top-3 features were used in the online experiment.}
\label{tab:top_features}
\end{table}